\documentclass[conference]{IEEEtran}
\IEEEoverridecommandlockouts
\usepackage{graphicx}
\usepackage{amsmath,amsfonts,amssymb}
\usepackage[caption=false]{subfig}
\usepackage{bbm}
\usepackage{cite}

\begin{document}
\bstctlcite{IEEEexample:BSTcontrol}

\title{%
Quantum Transfer Learning for Wi-Fi Sensing}
              
\author{\IEEEauthorblockN{Toshiaki Koike-Akino, Pu Wang, Ye Wang}
\IEEEauthorblockA{
\textit{Mitsubishi Electric Research Laboratories (MERL), Cambridge, MA 02139, USA}
\\
Email: \{koike, pwang, yewang\}@merl.com
}
}

\maketitle

\begin{abstract}
Beyond data communications, commercial-off-the-shelf Wi-Fi devices can be used to monitor human activities, track device locomotion, and sense the ambient environment. 
In particular, spatial beam attributes that are inherently available in the 60-GHz IEEE 802.11ad/ay standards have shown to be effective in terms of overhead and channel measurement granularity for these indoor sensing tasks. 
In this paper, we investigate transfer learning to mitigate domain shift in human monitoring tasks when Wi-Fi settings and environments change over time. 
As a proof-of-concept study, we consider quantum neural networks (QNN) as well as classical deep neural networks (DNN) for the future quantum-ready society.
The effectiveness of both DNN and QNN is validated by an in-house experiment for human pose recognition, achieving greater than 90\% accuracy with a limited data size. 
\end{abstract}


\section{INTRODUCTION}

Wi-Fi-based human activity monitoring has received much attention over the past decade due to the decreasing cost and less privacy concerns compared with camera-based approaches. 
Modern deep neural networks (DNNs) have made commercial Wi-Fi-band signals useful for human sensing tasks such as user identification, gesture recognition, device-free localization, fall detection, emotion sensing, and skeleton tracking~\cite{AdibHsu15, HsuLiu17, ZhaoLi18, ZhaoTian18, ZhaoLu19, SinghSandha19, LuWen16, ZengPathak16, WuZhang17, ZouZhou18, YangZou18, GuLiu18, GuZhang19, WangJiang19, WangZhou19, WangFeng19, ma2019wifi, ZhangRuan20, JiangXue20, PajovicWang19, WangPajovic19, KoikeWang20,WangKoike20, YuWang20}.
Nonetheless, Wi-Fi sensing is susceptible to time-varying fading, shadowing, path loss, hardware impairments (such as  carrier frequency offset, symbol timing offset, and sampling frequency offset), interference, and noise. 
Accordingly, DNNs designed at a training session may not be reliable enough at a time of testing.

In this paper, we investigate transfer learning (TL) or domain adaptation (DA)~\cite{wang2018deep} to be robust against such time-varying domain shifts for Wi-Fi sensing.
We experimentally validate the benefit of TL for DNNs to accurately recognize human's pose using commercial-off-the-shelf (COTS) Wi-Fi devices.
Besides standard DNNs, we introduce a new framework of quantum neural networks (QNNs) which leverage a quantum processing unit (QPU) as an alternative solution for an envisioned future era of quantum supremacy~\cite{arute2019quantum, zhong2020quantum}.
While quantum machine learning (QML) is considered as a potential driver in the sixth generation (6G) applications~\cite{nawaz2019quantum}, there are few research yet to tackle practical problems.
To the best of our knowledge, this is the very first paper studying QML for Wi-Fi sensing.
We specifically exploit a similar technique of quantum transfer learning (QTL) framework~\cite{mari2020transfer}.

Quantum computers have the potential to realize computationally efficient signal processing compared to traditional digital computers by exploiting quantum mechanism, e.g., superposition and entanglement, in terms of not only execution time but also energy consumption.
In the past few years, several companies including IBM, Google, and Honeywell have manufactured commercial quantum computers. 
For instance, IBM has released $127$-qubit QPUs available to the public via a cloud service.
Some groups reported to have achieved \emph{quantum supremacy} for specific problems~\cite{arute2019quantum, zhong2020quantum}.
It may be no longer far future when noisy intermediate-scale quantum (NISQ) computers~\cite{bharti2021noisy} will be widely used for various real applications.
Although quantum-ready algorithms for wireless communication systems have been investigated~\cite{babar2014exit, botsinis2015iterative, botsinis2016joint, botsinis2018quantum}, most existing works assume \emph{fault-tolerant} QPUs, which may be beyond the capability of near-term NISQ devices. 
Recently, hybrid quantum-classical algorithms based on the variational principle were proposed to achieve high robustness against quantum errors~\cite{farhi2014quantum, farhi2016quantum, anschuetz2019variational, kandala2017hardware}.
Hence, such modern variational quantum algorithms have not been studied for communications problem until very recently~\cite{matsumine2019channel, koike2020variational}.

We use variational quantum circuit (VQC)~\cite{schuld2020circuit} to build QNN~\cite{farhi2018classification} as a state-of-the-art QML approach~\cite{henderson2020quanvolutional, romero2017quantum, rebentrost2014quantum, lloyd2018quantum, dallaire2018quantum, verdon2019quantum}. 
The contributions of this paper are three-fold as described below:
\begin{itemize}
    \item This paper is the very first one to introduce QML for Wi-Fi sensing.
    \item We verify its feasibility for the  human pose recognition application with COTS Wi-Fi devices. 
    \item The TL framework is validated to improve the robustness against domain shifts across Wi-Fi scanning sessions.
\end{itemize}

\section{Wi-Fi Sensing for Human Monitoring}
\label{sec:algorithm}

For Wi-Fi sensing, we collect beam scanning measurements associated with a class of human gestures as a fingerprinting data to learn DNN and QNN models.

\begin{figure}
\centering
 \subfloat[][Wi-Fi pose recognition empowered by QML]{
 \includegraphics[width=0.9\linewidth]{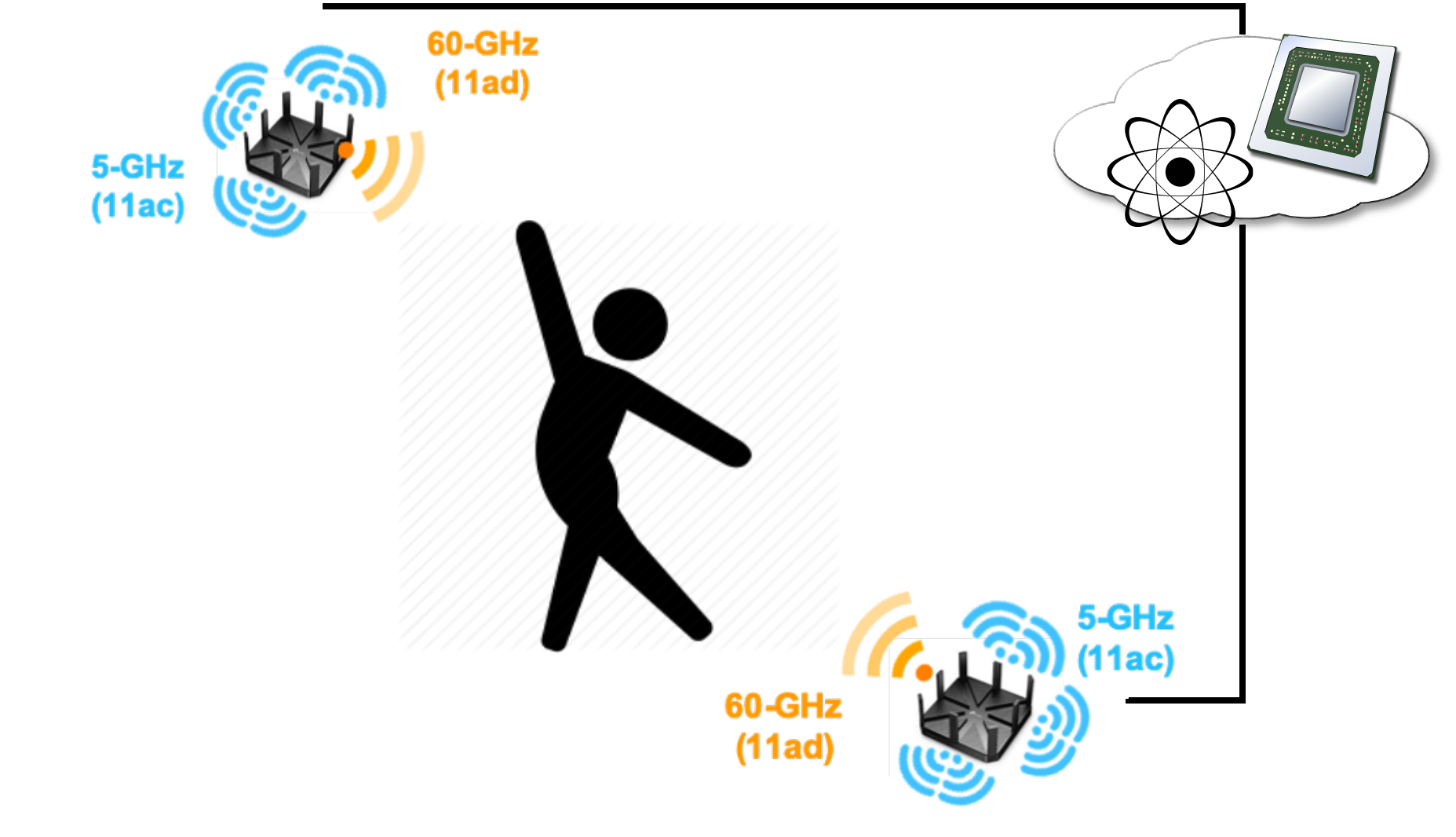}
 }\\
\subfloat[][Pose snapshots]{\includegraphics[width=\linewidth]{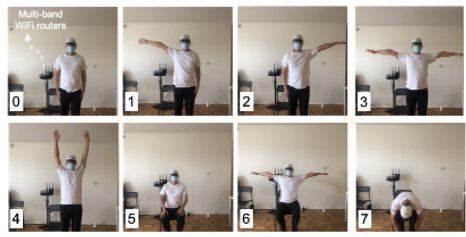}} 
\caption{Wi-Fi sensing tasks and experiment settings.  }
\label{fig:tasks}
\end{figure}

\subsection{Human Pose Recognition Experiments}

Fig.~\ref{fig:tasks} (a) shows our experimental configuration using one Wi-Fi station in front of a subject and another station behind the subject. 
Both stations are placed on a stand of a height of $1.2$~meters with a distance of approximately $2$ meters. 
As shown in Fig.~\ref{fig:tasks} (b), the subject is asked to perform a total of $8$ poses including distinct gestures like `sit', `stand with left arm lifted', etc. 
For each pose, we recorded $7$ independent data sessions with different time durations and with sufficient time separation between consecutive two sessions. 
We use the measurements in the first four data sessions as the training data (referred to as `source domain') and those in the last three sessions as the test data (`target domain'). 

The total number of measurement samples is $42{,}915$ and $1{,}040$ in the source domain and target domain datasets, respectively. 
The pose-wise distribution is listed in Table~\ref{tab:sample}.
We further divide the dataset in each domain into two sub-sets: i) labeled for training an ML model and ii) unlabeled for evaluating prediction accuracy of the ML model. 
For simplicity, no data augmentation was used. 

\begin{table}[t]
\caption{Number of beam SNR samples for each pose} 
\centering %
\begin{tabular}{c  c  c } %
\hline%
Pose & Source Domain & Target Domain  \\ %
\hline %
0 &  434 & 151  \\
1 & 499 & 149 \\
2 & 325 & 173 \\
3 & 347 & 129 \\ 
4 & 238 & 88 \\
5 & 314 & 96 \\
6 & 272 & 119 \\
7 & 432 & 135 \\
\hline %
Total & 42,915 & 1,040 \\
\hline
\end{tabular}
\label{tab:sample} %
\end{table}

\subsection{COTS Wi-Fi Testbed: mmWave Beam SNR}
\begin{figure}[t]
 \centering
 \includegraphics[width=\linewidth, trim={0 0 0 0}, clip]{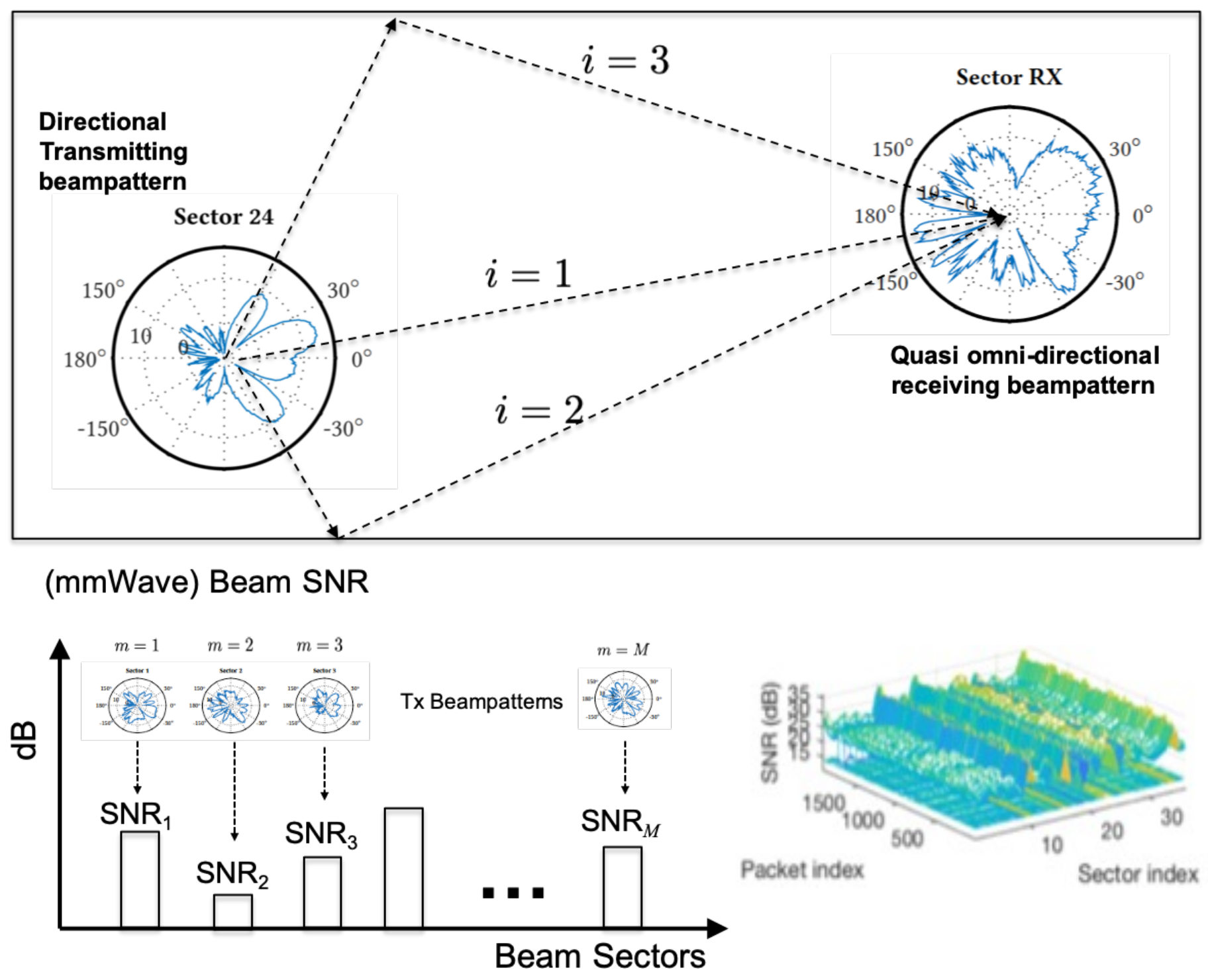}
\caption{Mid-grained beam SNR measurements at mmWave frequency bands, i.e., $60$ GHz in $802.11$ad standards.}
\label{fig:bSNR}	
\end{figure}

As super-grained mmWave channel state information is not accessible from COTS devices without additional overhead, we use \emph{mid-grained} Wi-Fi measurements in the beam angle domain---beam signal-to-noise ratios (SNRs)---generated from the beam training (a.k.a. beam alignment) phase. 
For each probing beampattern (a.k.a. beam sectors), beam SNR is collected by $802.11$ad devices as a measure of beam quality.  
Such beam training is periodically carried out and the beam sectors are adapted to environmental changes. 
Fig.~\ref{fig:bSNR} shows beam SNRs at pre-specified beampatterns used at the transmitting side. 
Access to raw mmWave beam SNR measurements from COTS devices is obtained 
via an open-source software~\cite{SteinmetzerWegemer18}. 

We use $802.11$ad-compliant TP-Link Talon AD7200 routers to collect beam SNRs at $60$ GHz.  
This router supports a single stream communication using analog beamforming over a $32$-element planar array. 
From one beam training, one Wi-Fi station can collect $36$ beam SNRs across discrete transmitting beampatterns. 
The measured beam SNRs are sent to a workstation via Ethernet cables to train DNN or QNN (e.g., through the IBM quantum cloud service). 
The experimental system is deployed in a standard indoor room setting. 
Further details of the experiments can be found in our previous work~\cite{YuWang20}

The key challenge of Wi-Fi sensing is that Wi-Fi measurements or the ambient environment may change over measurement sessions and these changes will degrade the efficiency of machine learning models due to the domain shifts.
We investigate a TL framework to tackle the domain shift issues, focusing on QTL~\cite{mari2020transfer} as described in the next section.

\begin{figure}[t]
\centering 
\includegraphics[width=\linewidth]
{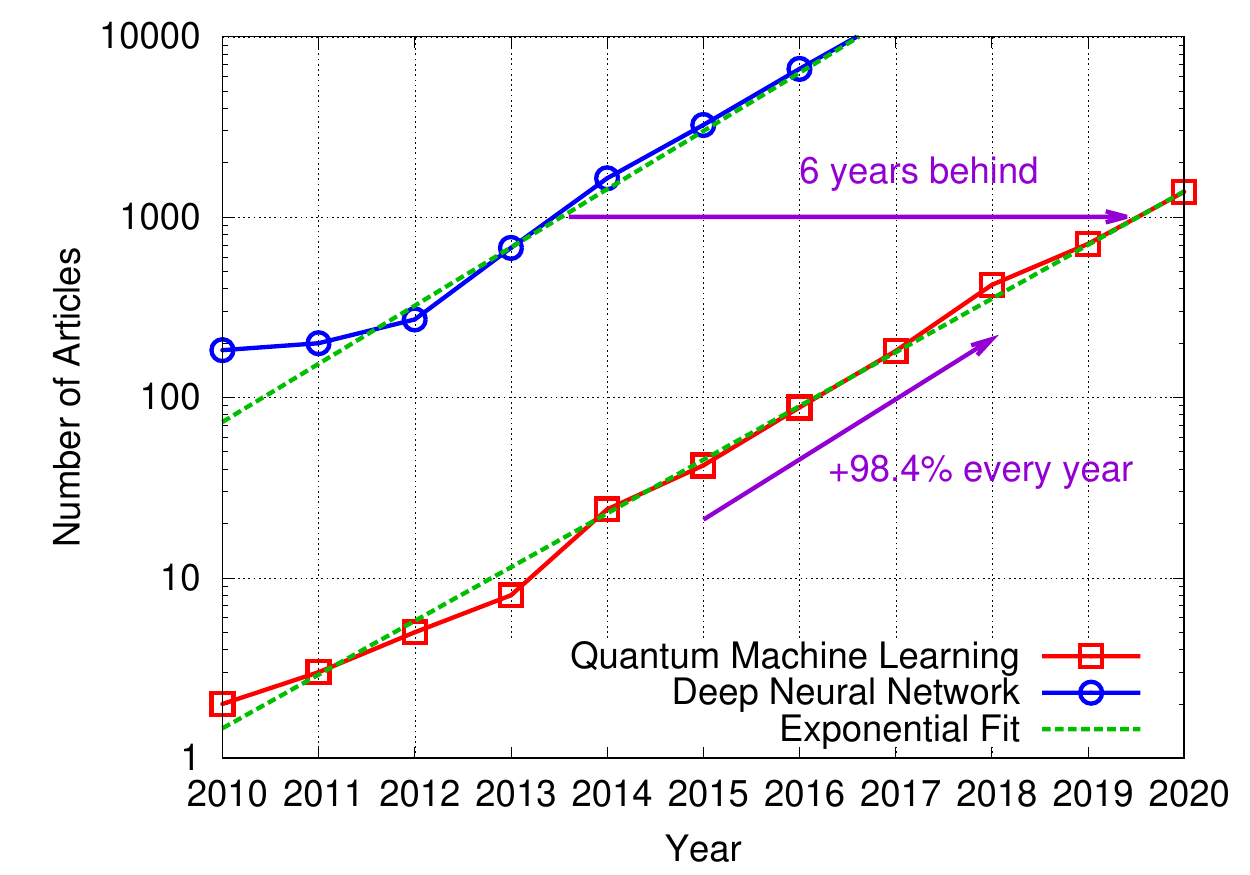}
\caption{Number of articles hit in Google scholar: articles on ``quantum machine learning'' has exponentially increased at an annual growth rate of 2 folds.
}
\label{fig:qml}
\end{figure}

\begin{figure*}[t]
\centering 
\includegraphics[width=0.95\linewidth]
{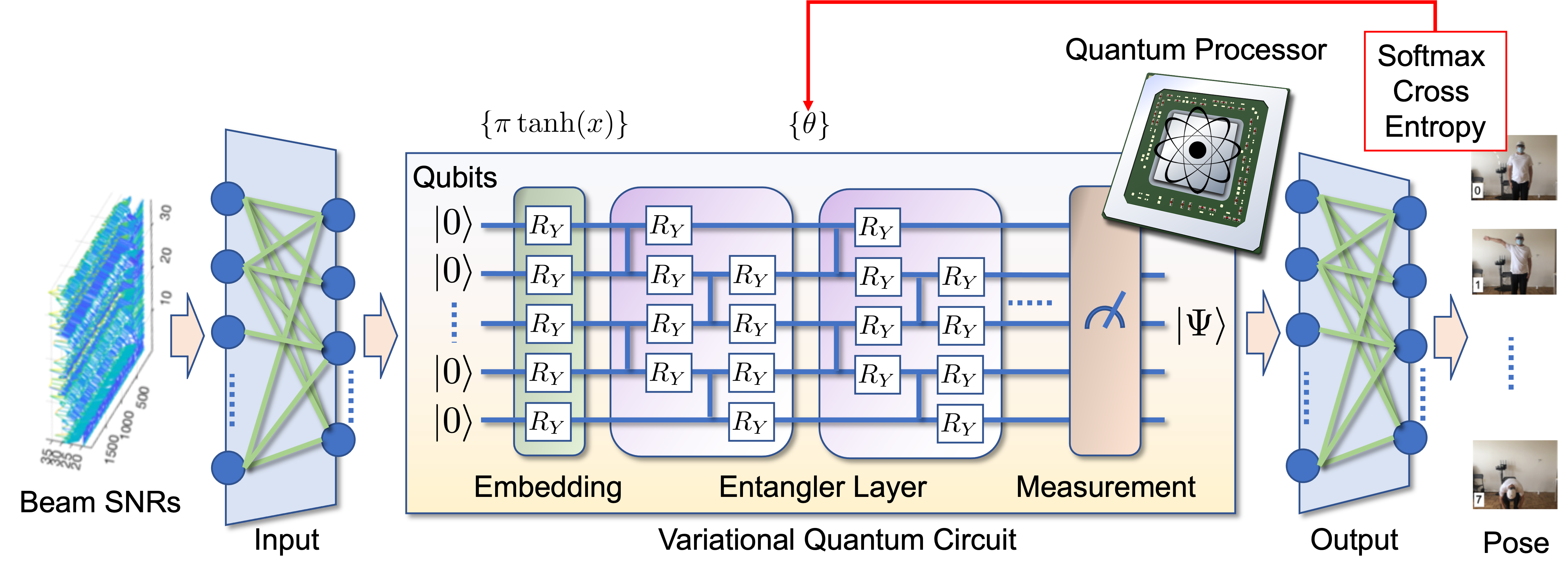}
\caption{Variational QNN using STD ansatz, dressed with input/output linear layers for pose recognition.
}
\label{fig:qnn}
\end{figure*}

\section{Quantum Transfer Learning (QTL)}

\subsection{Quantum Machine Learning (QML)}

Variational quantum algorithms~\cite{farhi2014quantum, farhi2016quantum, anschuetz2019variational, kandala2017hardware} are key enabling techniques for NISQ QPUs~\cite{bharti2021noisy}.
We previously demonstrated that they are useful for some communications problems in~\cite{matsumine2019channel, koike2020variational} by using real QPU computing at the IBM cloud service.
Variational quantum computing has been further integrated to machine learning, e.g., 
QNN~\cite{farhi2018classification}, quanvolutional network~\cite{henderson2020quanvolutional}, quantum autoencoder (QAE)~\cite{romero2017quantum}, quantum graph neural network (QGNN)~\cite{verdon2019quantum}, and quantum generative adversarial network (QGAN)~\cite{lloyd2018quantum, dallaire2018quantum}.

Fig.~\ref{fig:qml} shows a yearly trend of QML in the number of articles hit in Google scholar with keywords of `quantum machine learning' vs `deep neural network'. 
It is found that the number of QML articles has been exponentially increasing over the last decade with a growth rate of $2$ folds every year.
Interestingly, this is just behind the growth of DNN articles by six years.
It may potentially suggest that QML will be widely used in numerous research communities in a couple of years, considering the fact that applications of DNN are presently seen nearly everywhere.  

It was proved that QNN holds the universal approximation property~\cite{perez2020data}, which shows the close relation between the number of qubits in QNN and the number of DNN layers, and also the number of quantum layers of QNN and the number of hidden nodes in DNNs. 
Accordingly, increasing the number of qubits may enjoy state-of-the-art DNN performance.
More importantly, quantum circuits can be analytically differentiable with a parameter-shift rule~\cite{schuld2019evaluating} that enables stochastic gradient optimization of VQC.
Nevertheless, QNN often suffers from a vanishing gradient issue called the barren plateau~\cite{mcclean2018barren}.
To mitigate the issue, some techniques have been studied, e.g, an identity initialization strategy~\cite{grant2019initialization} and a simplified two-design (STD) ansatz~\cite{cerezo2021cost} to cover nearly arbitrary unitary spectrum with shallow staggered entanglers.

\subsection{Quantum Supremacy}

Although quantum supremacy over classical computers has been reported for some specific problems~\cite{arute2019quantum, zhong2020quantum}, the feasibility of realizing quantum advantages for practical problems has still been argued intensively. 
Nevertheless, it is highly expected that quantum computers could provide breakthroughs in a wide range of research fields. 
This is particularly due to the known limits of power efficiency for classical computers, especially for deep learning, where training has become extremely energy intensive~\cite{strubell2019energy}.
For sustainable growth, a new computing modality such as quantum computers is demanded for a future eco-friendly society.
Although no one can certainly foresee that QPU will supersede classical computers in future, it is of importance to explore many possibilities. 

\subsection{Quantum Neural Network (QNN) Transfer Learning}

Fig.~\ref{fig:qnn} depicts QNN using STD ansatz~\cite{cerezo2021cost}, which consists of Pauli-Y rotations and staggered controlled-Z entanglers in sequence.
The STD ansatz is a simplified variant of a $2$-design whose statistical properties are identical to ensemble random unitaries with respect to the Haar measure up to the first $2$ moments.
For an $n$-qubit VQC, there are $2(n-1)L$ variational parameters $\{\theta\}$ over an $L$-layer STD ansatz.
To embed $36$-dimensional beam SNRs, an input linear layer is used to initialize the quantum state for rotation angles of Pauli-Y gates.
The $8$-class pose estimation is provided by quantum measurements in the Hamiltonian observable of Pauli-Z operations, followed by an output layer to align the dimension.
The variational parameters as well as input/output layers are optimized by the adaptive momentum stochastic gradient descent to minimize the softmax cross entropy loss.
The QNN model is first trained by labeled data in source domain, and then fine-tuned with few-shot labeled data from the target domain for transfer learning while input/output layers are frozen.
While QNN is not necessarily better than DNN in prediction accuracy, the potential advantage of QNN lies in its computational efficiency to manipulate $2^n$ quantum states at once with a small number of quantum gates.

\section{Performance Evaluation}
\label{sec:performance}

\subsection{Comparison of ML Methods}

We first evaluate the baseline performance in the human pose recognition task without any TL techniques: specifically, ML models are trained using only the labeled data in the source domain and tested in the target domain. 
Besides DNN and QNN, we compare different ML methods including support vector machine (SVM), decision tree (DT), $k$-nearest neighbor ($k$NN), Gaussian na\"{i}ve Bayes (GNB), random forest (RF), bagging ensemble, and extra tree ensemble methods ($10$ base models).
For DNN, we consider residual $4$ hidden layers with $100$ hidden nodes using Mish activation, where there are approximately $35$k trainable parameters.
For QNN, we use $10$ qubits with $1$-layer STD ansatz~\cite{cerezo2021cost}, dressed by input and output linear layers, where there are $18$ quantum variational parameters and $610$ classical trainable parameters in total.
We use AdamW optimizer for a mini batch size of $100$ over $100$ epochs with a learning rate of $0.02$ and weight decay of $10^{-4}$.
For simplicity, we consider no quantum noise for simulations.

\begin{figure}[t]
\centering 
\includegraphics[width=\linewidth]
{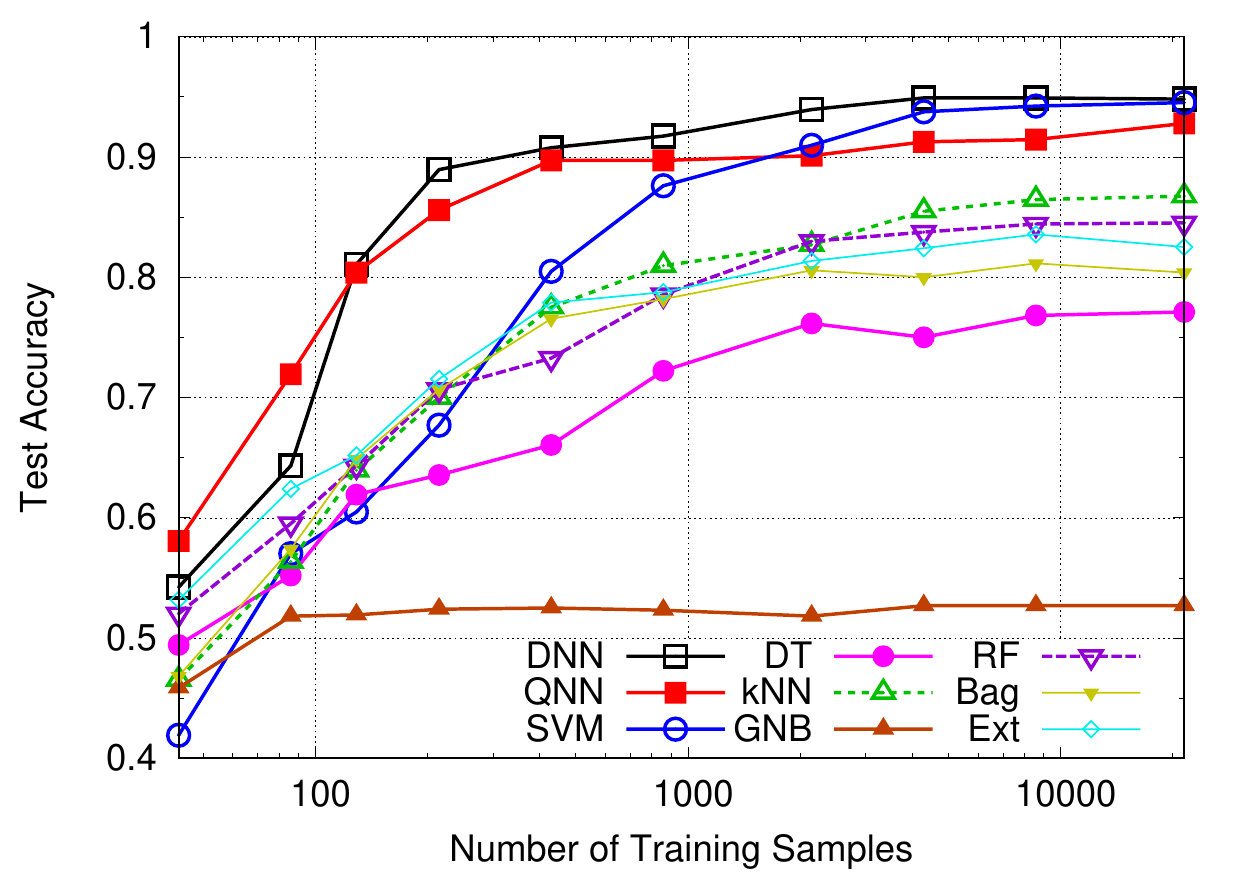}
\caption{Test accuracy (target domain) vs. training samples (source domain).}
\label{fig:pose_train}
\end{figure}

Fig.~\ref{fig:pose_train} shows the test accuracy as a function of the number of labeled training samples in the source domain.
One can notice that the performance can exceed an accuracy of $90\%$ for DNN, QNN, and SVM when a sufficient amount of labeled data is available.
It is confirmed that a small-scale QNN can achieve performance comparable to a large-scale DNN.
Most conventional ML techniques are found ineffective for our pose recognition task.

\begin{figure*}[t]
 \centering
 \subfloat[][SVM (mean acc: $60.5\%$)]{
 \includegraphics[width=0.23\linewidth, trim={50 0 60 0}, clip]{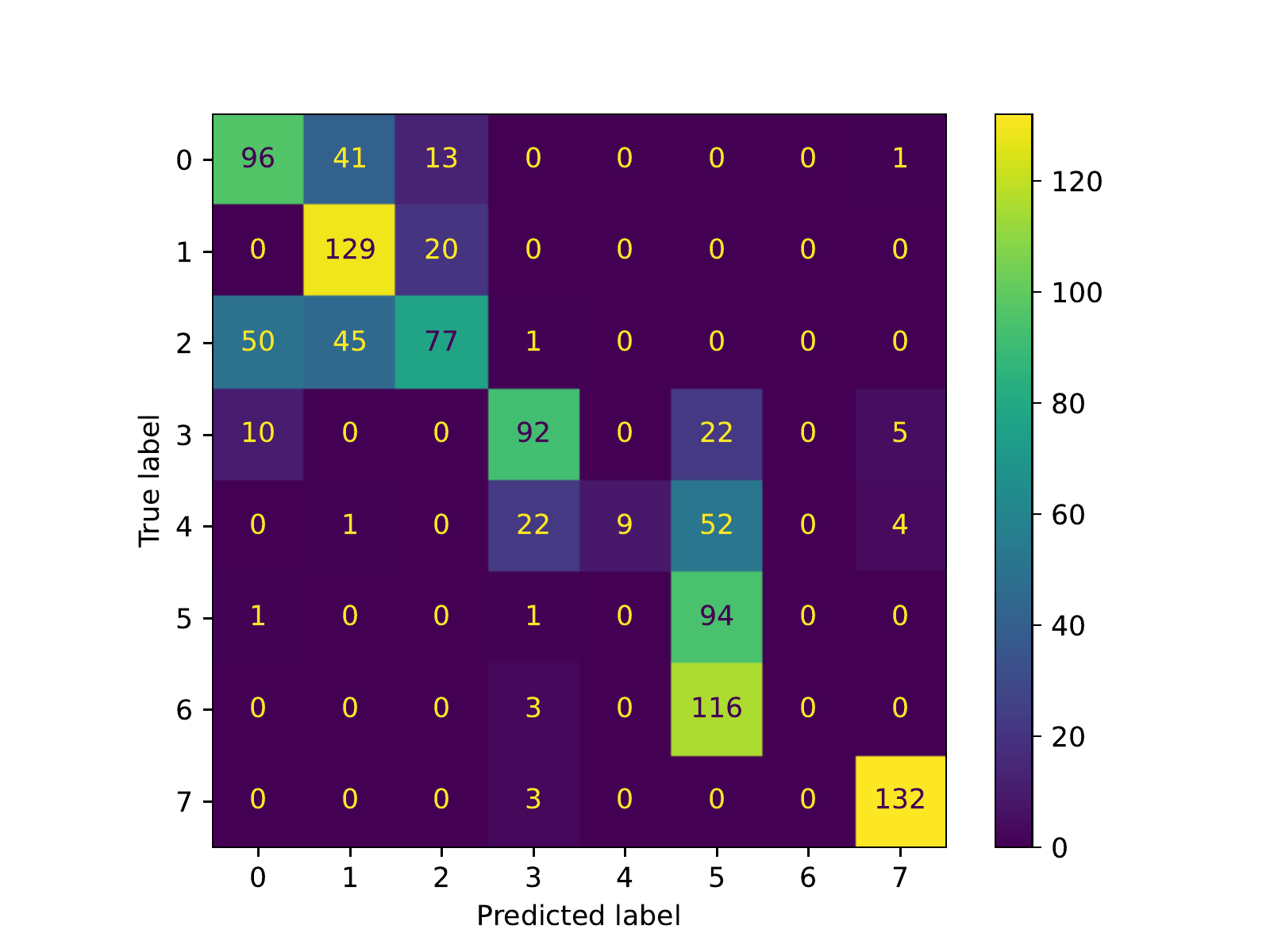}
 }
 \hfill
 \subfloat[][DNN (mean acc: $81.1\%$)]{
 \includegraphics[width=0.23\linewidth, trim={50 0 60 0}, clip]{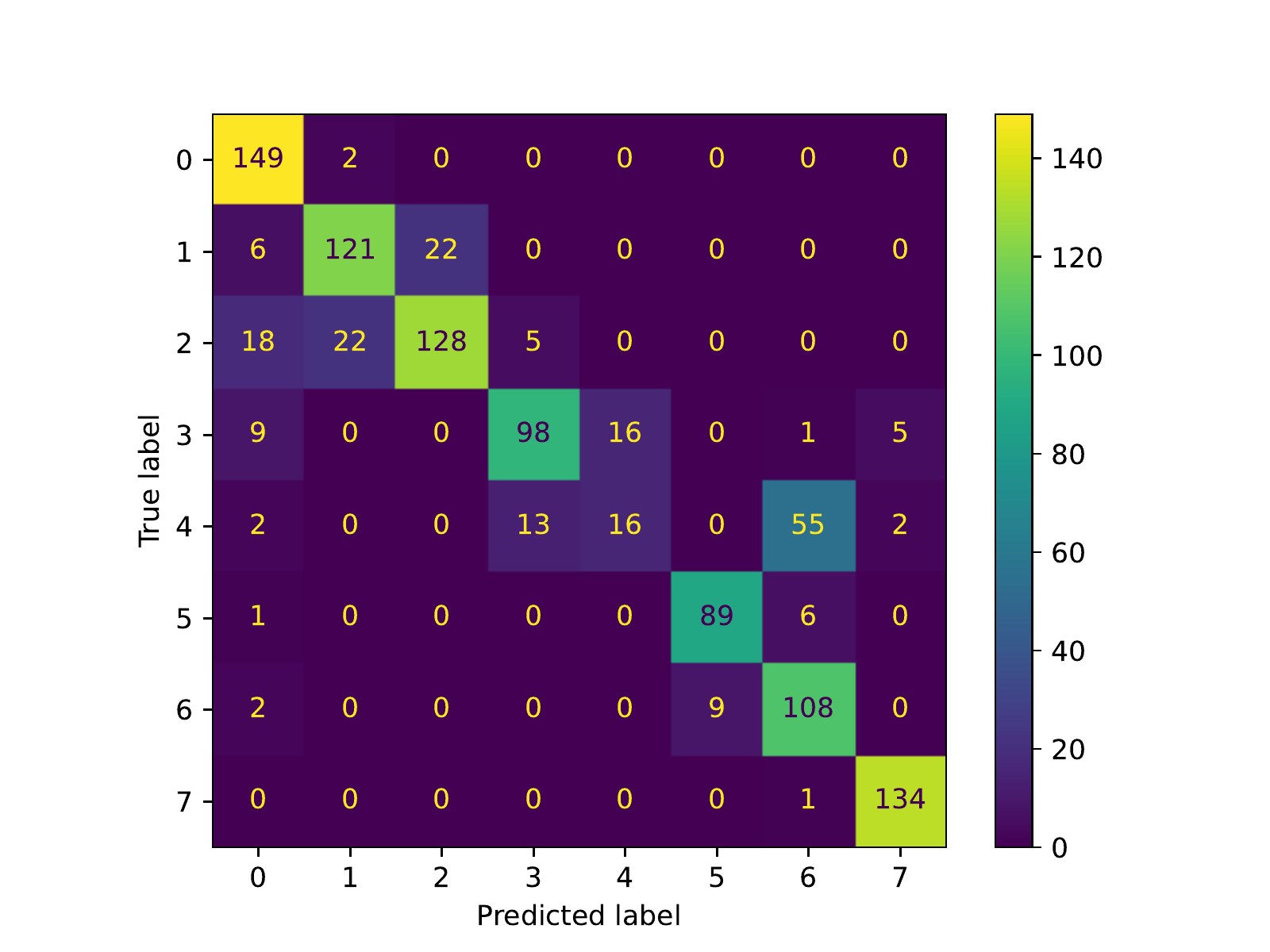}
 } 
 \hfill
 \subfloat[][QNN (mean acc: $80.4\%$)]{
 \includegraphics[width=0.23\linewidth, trim={50 0 60 0}, clip]{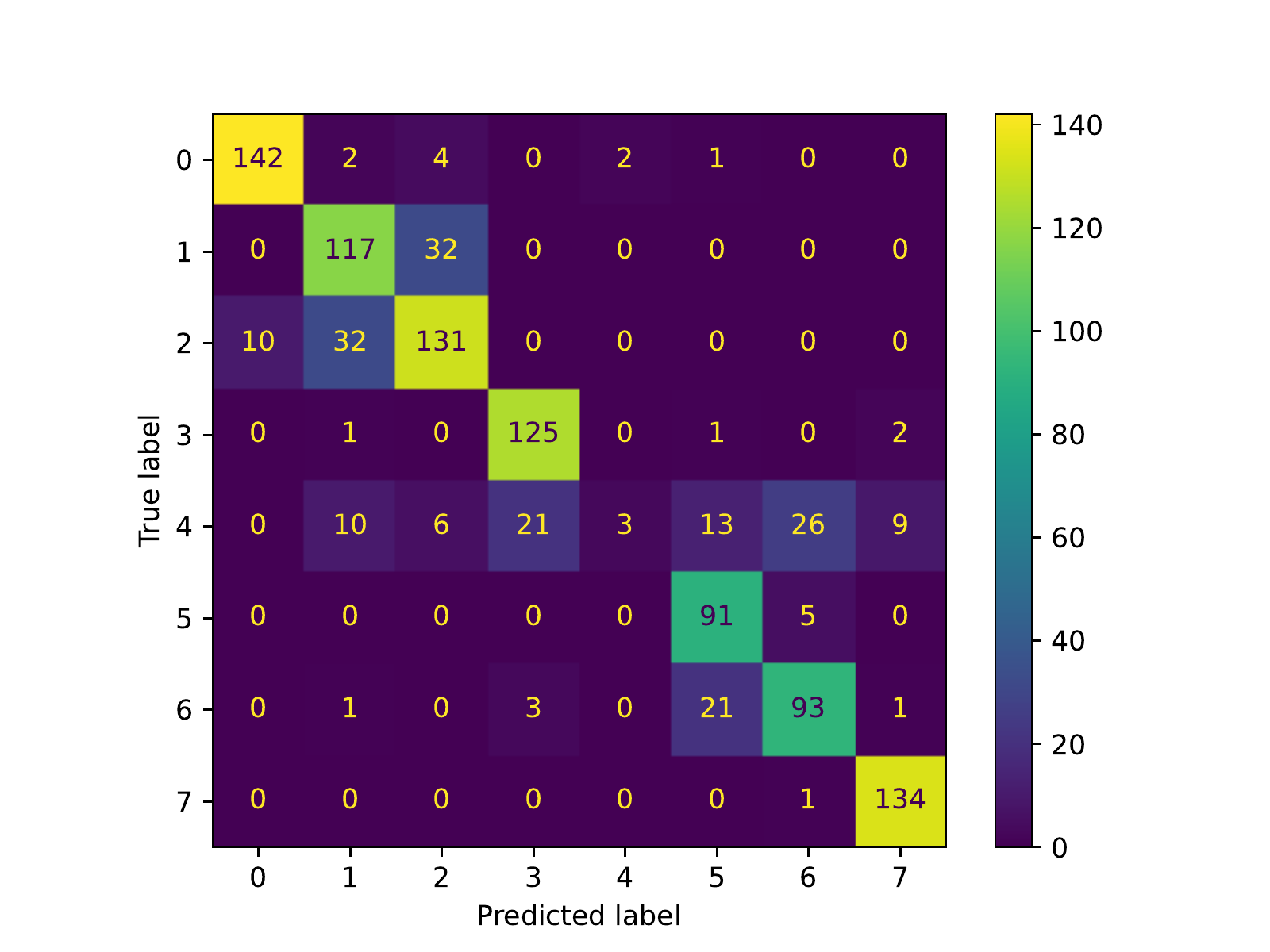}
 } 
 \hfill
 \subfloat[][TL-QNN (mean acc: $90.7\%$)]{
 \includegraphics[width=0.23\linewidth, trim={50 0 60 0}, clip]{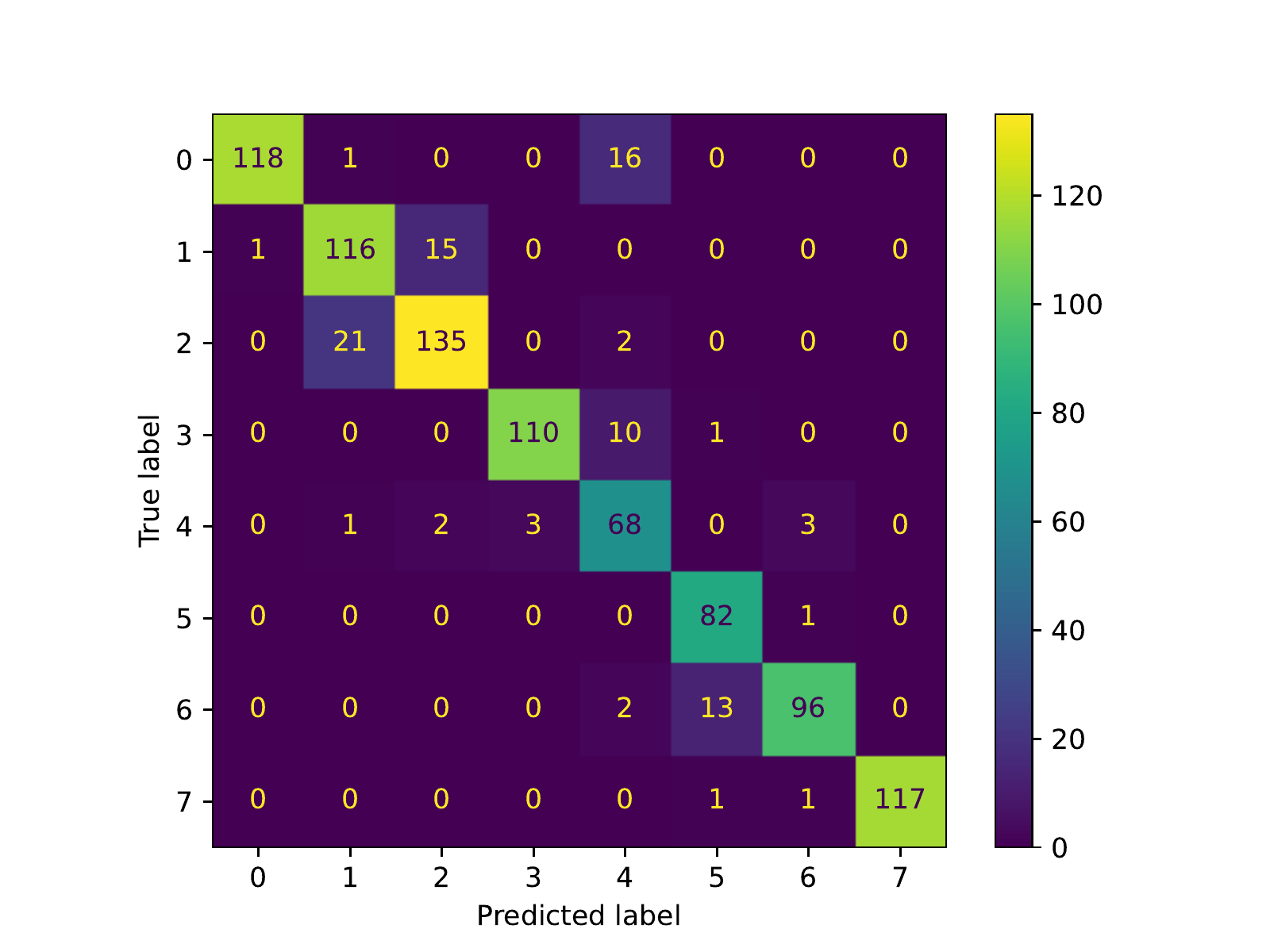}
 } 
 \\
 \subfloat[][SVM (AUC: $0.933$)]{
 \includegraphics[width=0.23\linewidth, trim={20 0 30 30}, clip]{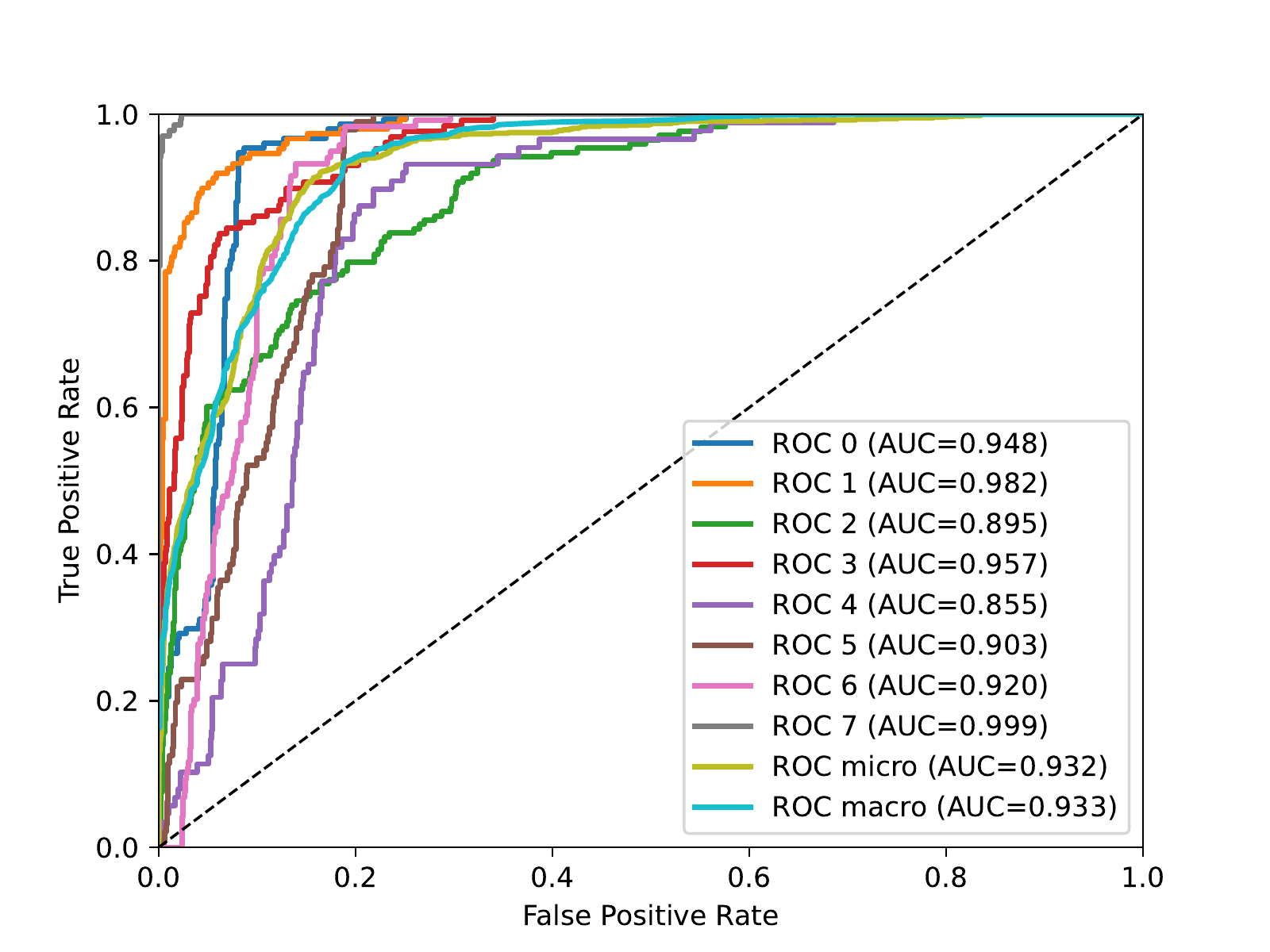}
 } 
 \hfill
 \subfloat[][DNN (AUC: $0.958$)]{
 \includegraphics[width=0.23\linewidth, trim={20 0 30 30}, clip]{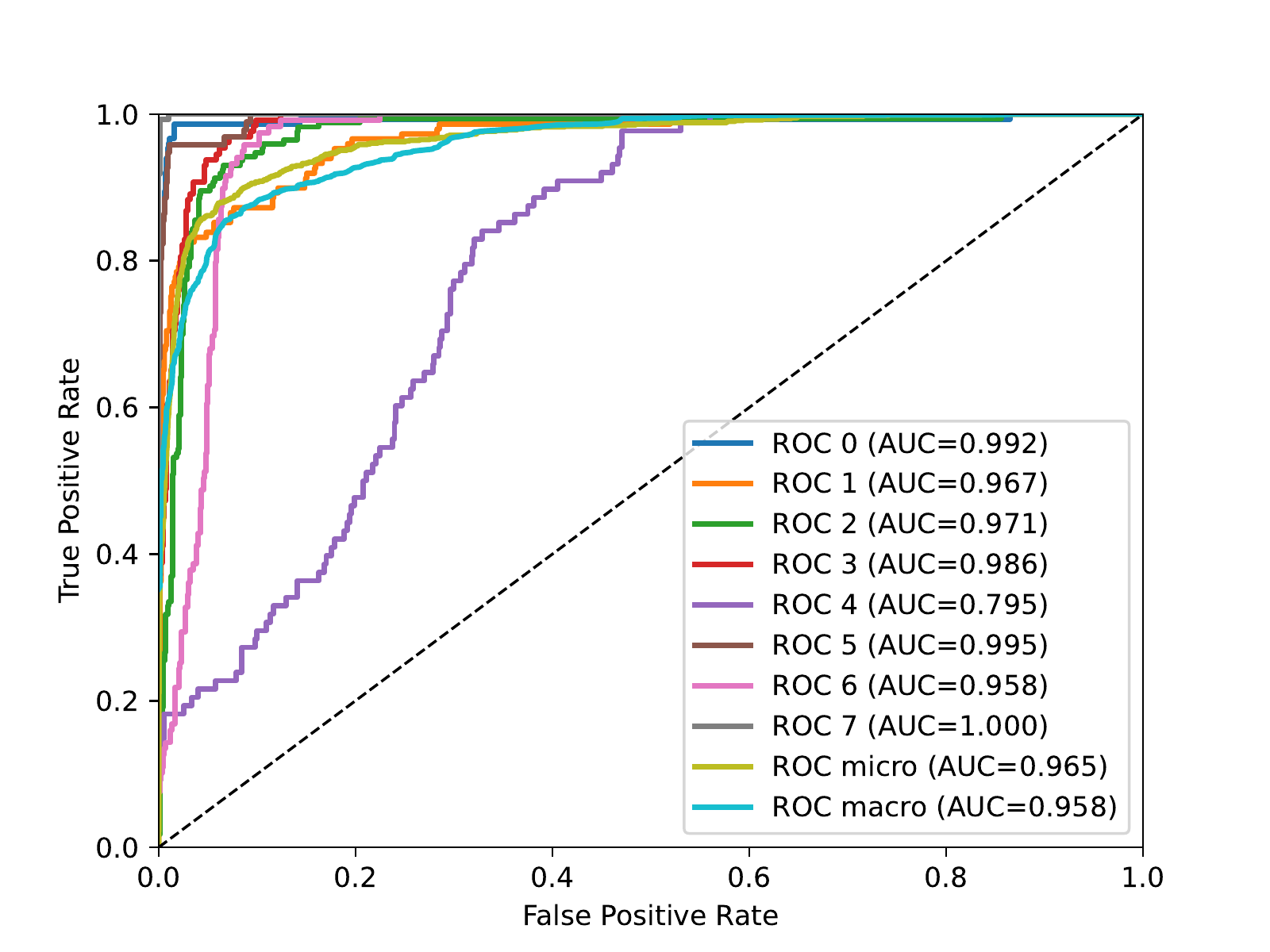}
 }
 \hfill
 \subfloat[][QNN (AUC: $0.971$)]{
 \includegraphics[width=0.23\linewidth, trim={20 0 30 30}, clip]{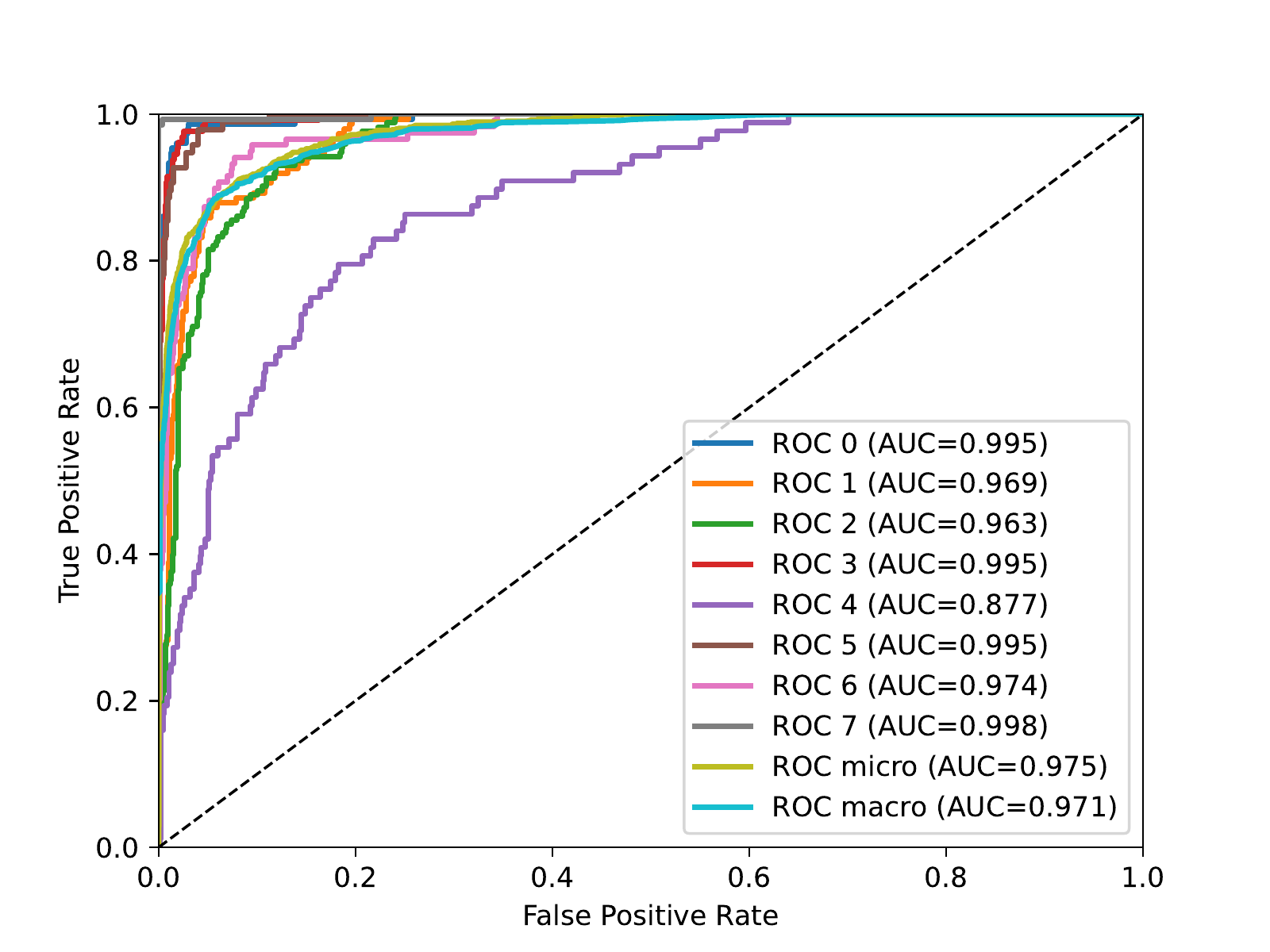}
 }
  \hfill
 \subfloat[][TL-QNN (AUC: $0.990$)]{
 \includegraphics[width=0.23\linewidth, trim={20 0 30 30}, clip]{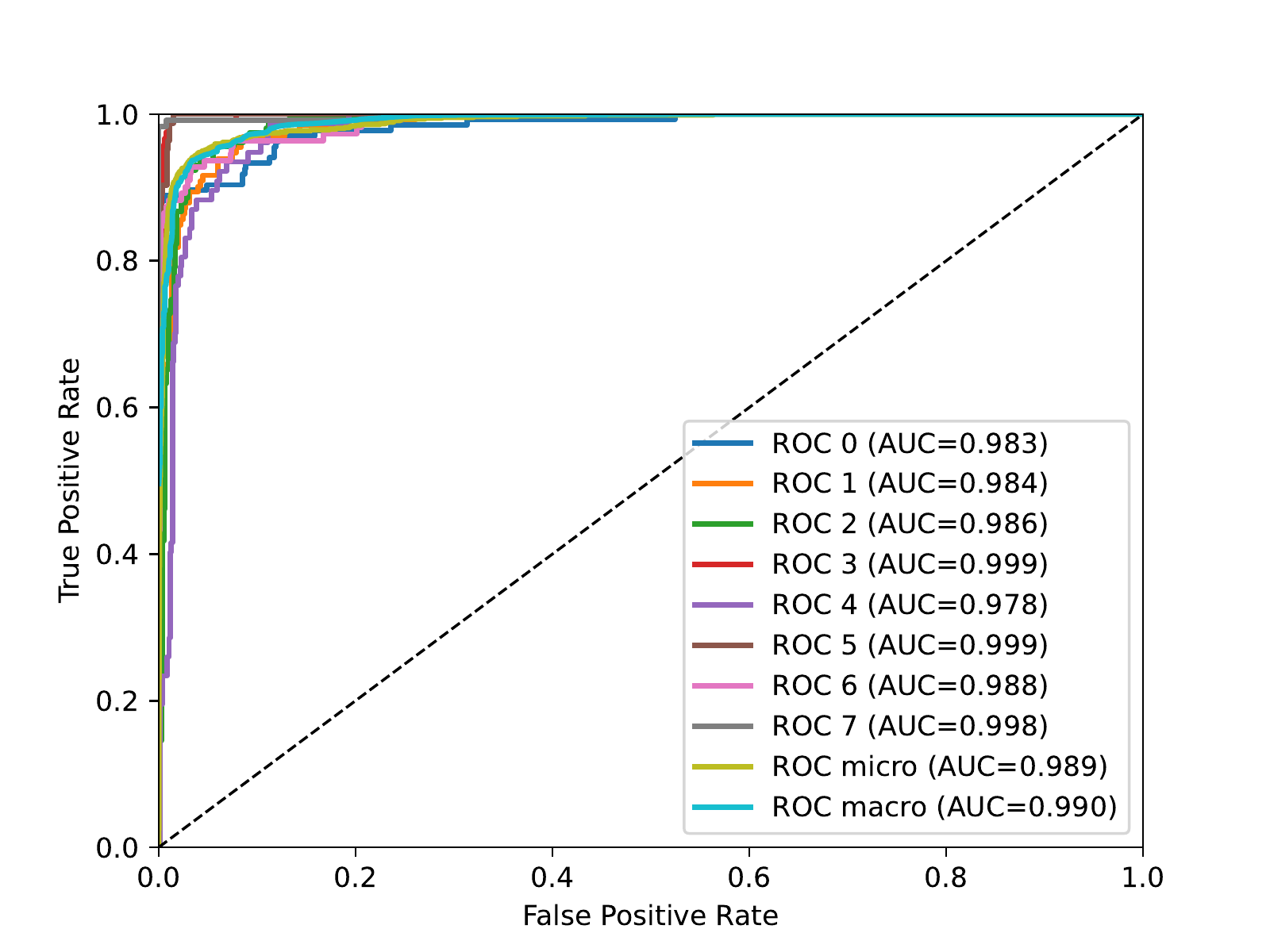}
 }
\caption{Confusion matrices (top row) and ROC curves (bottom row) for $8$-pose recognition with $129$ training samples ($0.3\%$ data labeled in the source domain). 
TL-QNN uses $104$ transfer samples ($10\%$ data labeled in the target domain).}
\label{fig:cm}	
\end{figure*}

\subsection{Pose Recognition Analysis}

We further discuss pose-wise detection performance in terms of confusion matrix and receiver operating characteristic (ROC) curve in Fig.~\ref{fig:cm}.
Here, we present performance of SVM, DNN, and QNN when $104$ training samples ($0.03\%$ data labeled in the source domain) are available to learn those models.
From the confusion matrices, the prediction of pose class $4$ was found to be the most challenging. 
It is particularly significant for DNN models in terms of ROC area under curve (AUC) score as shown in Fig.~\ref{fig:cm}(f) for the pose class $4$.
Although the mean accuracy of DNN was slightly better than that of QNN, QNN outperformed DNN in the sense of macro/micro-averaging ROC-AUC performance. 

\subsection{Domain Shift Impact}

\begin{figure}[t]
\centering 
\includegraphics[width=\linewidth]
{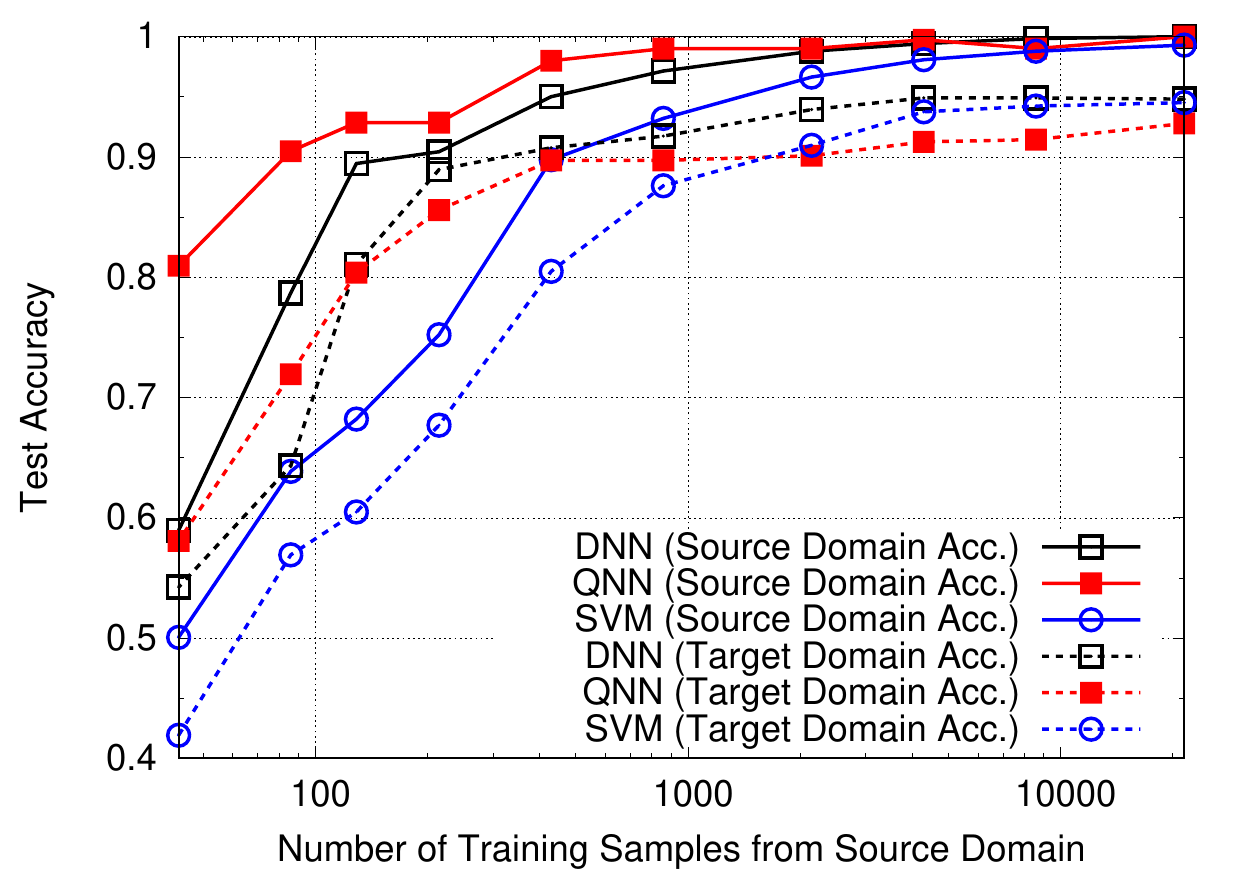}
\caption{Domain shift analysis with test accuracy (source/target domains) vs. training samples (source domain).}
\label{fig:pose_domain}
\end{figure}

Although DNN/QNN/SVM showed a reasonable prediction with an accuracy greater than $90\%$ in Fig.~\ref{fig:pose_train} for a sufficient size of training data in the source domain,
it does not exceed $95\%$.
This is mainly because of the domain shift issues across measurement sessions as shown in Fig.~\ref{fig:pose_domain}, where we present the test accuracy of unlabeled source domain data when training with labeled source domain data.
All ML models could achieve nearly perfect accuracy if the test data comes form the same domain.
In particular, QNN was significantly better than DNN.

\subsection{Transfer Learning}

\begin{figure}[t]
\centering 
\includegraphics[width=\linewidth]
{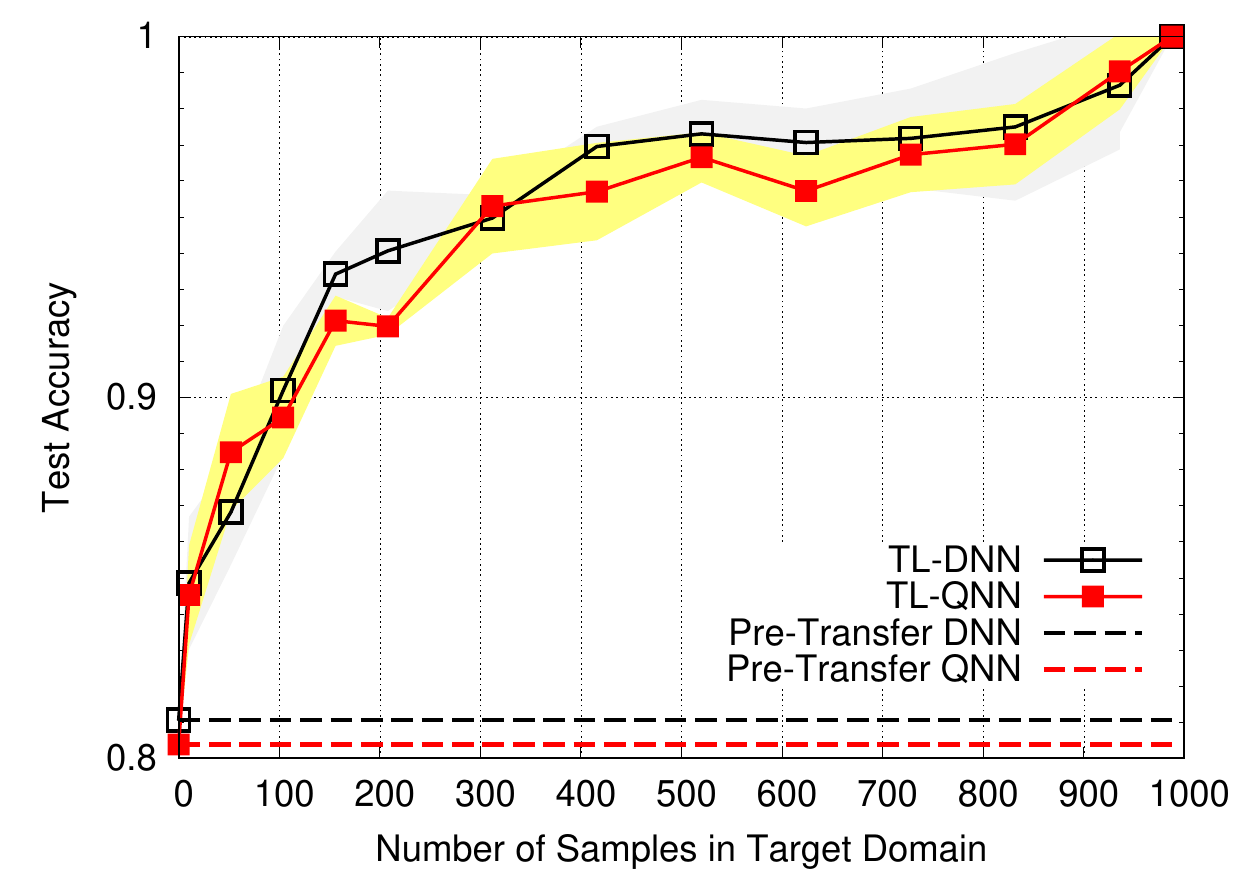}
\caption{Test accuracy vs. transfer samples (labeled data in target domain). 
Pre-transfer model uses $129$ training samples ($0.3\%$ data labeled in source domain).}
\label{fig:pose_trans}
\end{figure}

Fig.~\ref{fig:pose_trans} shows TL performance for DNN and QNN using few-shot transfer samples from the target domain.
Here, the pre-transfer model uses $129$ training samples from the source domain. 
We present mean accuracy with $1$-standard deviation zones over $5$-times fine tuning.
It is confirmed that TL-QNN is comparable to TL-DNN, achieving greater than $95\%$ accuracy over $300$ samples.
The corresponding confusion matrix and ROC curves for TL-QNN at $104$ transfer samples are shown in Fig.~\ref{fig:cm}(d) and (h), respectively.
We can observe that the TL can predict the pose class $4$ accurately with a significant improvement in ROC-AUC.

\section{Conclusion}
\label{sec:conclusion}
This paper considered transfer learning of DNN and QNN to deal with domain shifts due to changes of Wi-Fi measurements and environments over data sessions for Wi-Fi sensing tasks.
Specifically, we demonstrated the benefit of TL through real-world experiments with an in-house Wi-Fi testbed. 
We showed a small-scale QNN can achieve greater than $90\%$ accuracy, comparable to a large-scale DNN, in human pose recognition. 
Demonstration on real quantum processors will be provided in a future work.
This is a very initial proof-of-concept study for quantum-ready Wi-Fi sensing and there remain many fascinating open issues to solve in future.

\section*{Acknowledgment}
We thank Jianyuan Yu (Virginia Tech.) for earlier data collection experiments.

\bibliographystyle{IEEEtran}
\bibliography{refs}

\end{document}